\documentclass[letterpaper, 10 pt, conference]{ieeeconf}  

\IEEEoverridecommandlockouts                              

\usepackage{graphics} 
\usepackage{amsmath} 
\usepackage{amssymb}  
\usepackage{amsfonts} 
\usepackage{subcaption}
\usepackage{graphicx} 
\usepackage{algorithm}
\usepackage{algpseudocode}

\DeclareMathOperator*{\argmin}{arg\,min}
\makeatletter
\newenvironment{breakablealgorithm}
  {
   \begin{center}
     \refstepcounter{algorithm}
     \hrule height.8pt depth0pt \kern2pt
     \renewcommand{\caption}[2][\relax]{
       {\raggedright\textbf{\ALG@name~\thealgorithm} ##2\par}%
       \ifx\relax##1\relax 
         \addcontentsline{loa}{algorithm}{\protect\numberline{\thealgorithm}##2}%
       \else 
         \addcontentsline{loa}{algorithm}{\protect\numberline{\thealgorithm}##1}%
       \fi
       \kern2pt\hrule\kern2pt
     }
  }{
     \kern2pt\hrule\relax
   \end{center}
  }

\title{\LARGE \bf
Efficient State Estimation with Constrained Rao-Blackwellized Particle Filter
}

\author{Shuai Li$^{1}$, Siwei Lyu$^{2}$ and Jeff Trinkle$^{3}$
\thanks{$^{1}$Shuai Li is with GE Global Research Center, 1 Research Circle, Niskayuna, NY, 12309, USA
        {\tt\small lis12@rpi.edu}}%
\thanks{$^{2}$Siwei Lyu is with Faculty of Computer Science,  University at Albany, SUNY, Albany, NY 12222, USA
        {\tt\small lsw@cs.albany.edu}}%
        \thanks{$^{3}$Jeff Trinkle is with Faculty of Computer Science,  Rensselaer Polytechnic Institute, Troy, NY 12180, USA
        {\tt\small trink@cs.rpi.edu}}%
}

\begin{document}

\maketitle
\thispagestyle{empty}
\pagestyle{empty}

\begin{abstract}
Due to the limitations of the robotic sensors, during a robotic manipulation task, the acquisition of the object's state can be unreliable and noisy. Combining an accurate model of multi-body dynamic system with Bayesian filtering methods has been shown to be able to filter out noise from the object's observed states.  However, efficiency of these filtering methods suffers from samples that violate the physical constraints, e.g., no penetration constraint. 

In this paper, we propose a Rao-Blackwellized Particle Filter (RBPF) that samples the contact states and updates the object's poses using Kalman filters. This RBPF also enforces the physical constraints on the samples by solving a quadratic programming problem. By comparing our method with methods that does not consider physical constraints, we show  that our proposed RBPF is not only able to estimate the object's states, e.g., poses, more accurately but also able to infer unobserved states, e.g., velocities, with higher precision.
\end{abstract}

\section{Introduction}
To perform a robotic manipulation task, such as moving a cup to a certain location on top of a table, interactions between the robot and the object are inevitable. As a result, the robot together with the objects that it manipulates forms a multi-body dynamic system. In this dynamic system, the robot constantly adjusts its actions based on its observed states (in the context of robotic manipulation, including  the pose,  the velocity and the contact state ) of the objects and thus the accuracy of the objects' states are critical for the success of a robotic manipulation task. However, because of the limitations of the robotic sensors, the acquisition of the objects' poses and velocities can be unreliable and noisy, and some  properties of the contact state are difficult to be obtained, such as the sticking/sliding status.

To handle the state measurement issues for robotic manipulation tasks, we propose to incorporate a model of multi-body dynamics to both reduce measurement noise and infer unobserved states, such as the velocities.  The Linear Complementarity Problem (LCP) model is an accurate model of multi-body dynamics. In the LCP model, the motions and the contact states are formulated by the Newton-Euler equations and a set of complementarity constraints. Additionally, the LCP model is a piece-wise linear model. Previously, we developed the contact-based Rao-Blackwellized Particle Filter (RBPF) that estimates the pose and the contact state of the objects simultaneously by exploiting the piece-wise linear property of the LCP model \cite{shuai-icra-2015}. In the contact-based RBPF, the distribution of the contact state is estimated with particles. A set of equality constraints are derived with the estimated contact state, which are then used to convert the LCP model to a linear model. The motions and the poses of the objects are estimated with the linear model using Kalman filters \cite{Kalman-Filter}. 

\begin{figure} [h!]
\centering
\begin{subfigure}{.45\linewidth}
  \centering
  \includegraphics[width=.9\linewidth]{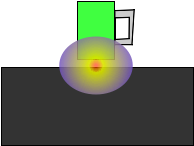}
  \caption{Estimated state distribution without enforcing the contact constraints.}
  \label{fig:unconstrained-example-motivation}
\end{subfigure}%
\hfill 
\begin{subfigure}{.45\linewidth}
  \centering
  \includegraphics[width=.9\linewidth]{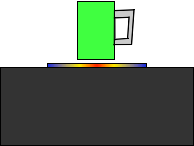}
 \caption{Estimated state distribution with enforced contact constraints.}
   \label{fig:constrained-example-motivation}
\end{subfigure}
\caption{Estimated state distributions of a cup that is about to be in contact with a rigid surface.}
\label{fig:example-motivation}
\end{figure}

For a multi-body dynamic system consists of rigid bodies, contacts imply strict constraints on the motions of the bodies. These constraints include no penetrations between two bodies and contact forces can only be repulsive. Ignoring these contact constraints causes two problems: (a) the noisy measurements can contaminate the estimation state distribution leading to even larger coverage of the invalid state space. (b) the probability density mass for the invalid state space is wasted even if it contains information that helps improving the estimation accuracy in the unconstrained directions. To elaborate, Figure \ref{fig:unconstrained-example-motivation} shows an example of a state estimation problem, where we want to estimate the position of the cup that is about to be in contact with a rigid surface. Without considering the contact constraints, although the mean of the estimated distribution shows the correct position of the cup,  a large portion of probability mass representing the cup penetrating the surface. On the other hand, if the contact constraints are enforced, the distribution should be projected and collapse to a one-dimensional distribution on the rigid surface as shown in Figure \ref{fig:constrained-example-motivation}. To tackle this problem and leverage the constraints, we propose the extend our previous contact-based RBPF and incorporate the equality constraints in the Kalman filter updating step by solving a quadratic programming problem. From our experiments, we show that this constrained contact-based RBPF can not only improve the performance of estimating the poses of the objects but also increase the accuracy of inferring the objects' unmearued velocities. 
\section{Related Work}   
To solve the state estimation problem for robotic manipulation tasks, some of the previous works developed algorithms that improve the accuracy of the estimation of the object's state by exploiting the features of the sensors. Hebert et al. tried to solve the state estimation problem of a rectangular block held in a robotic gripper \cite{hebert2011} by considering various sensor inputs as measurements including readings from a wrist force/torque sensor, finger joint position sensors, and stereo vision sensors. These measurements are used in an extended Kalman filter to estimate the pose of the block relative to the robotic gripper.  Meeussen et al. designed a hybrid filter by incorporating a graph of all possible contact states and their transitions \cite{Meeussen2006} to estimate the contact states and the geometric parameters. Their method tracks the contact states through a particle filter and updates the estimated probability distribution through a contact distance model and a residue measurement model. The work by Koval et al. developed the ``manifold particle filter" to support their work in planar push-grasping experiments \cite{Koval2015}. An interesting problem that can arise in contact tasks is that as contacts form, the dimension of the valid portion of the configuration space for the system drops, which can lead to particle starvation. The manifold particle filter solves this problem by leveraging the known positions of the contact sensors on a robotic hand. Their method samples particles on a pre-computed contact manifold, which includes the configuration space of the tracked object being in contact with the robotic hand. Chalon et al. developed a particle filter that tracks the pose of an object when it is grasped by a robot gripper \cite{Chalon2013}. Their method updates the pose of the object through the kinematic constraints of the robotic hand and incorporates both the hand joint sensors and the contact sensors in their measurement model.

The works above assume continuous measurements of the object and the motions of the object to be quasi-static. On the contrary, if we treat the objects and the robot as a multi-body dynamic system, a model of the multi-body dynamics can be used to infer the dynamic properties and the state during sensor occlusions. Many previous works together with our work fall into this category. Duff et al. combined a physics simulator (PhysX) with an RANSAC algorithm to estimate the trajectory of an object's motion \cite{duff2010}. The inputs to their algorithm are noisy position measurements of the object. The RANSAC algorithm selects inliers and outliers by fitting simulated trajectories to the observed ones. While our work solves the on-line state estimation problem, their method estimates the trajectory off-line. A follow-up work by Duff et al. used a real-time physics simulator (PhysX) as the basis of the dynamics model of a particle filter so as to track the poses of objects \cite{duff2011}. They demonstrated that this particle filter can track the pose of an object very well even during an occlusion. In their method, the simulator is used as a black box. Our proposed method, on the other hand, fully exploits a model of multi-body dynamics and converts the state estimation problem into a hybrid state estimation problem. Zhang et al. developed a general solution by combining a well-developed dynamic Bayesian network with a Rao-Blackwellization particle filter to speed up the state estimation of dynamic systems \cite{zhang2013dynamic}. Later, Zhang compared three different particle filters in a push grasping experiment and proposed a relaxed algorithm to break the nonlinear state transition model of the dynamic system into linear equations in a particle filtering implementation \cite{ZhangThesis}. The key difference between our approach and theirs is that instead of the continuous state space, our particle filter samples the contact states , which as shown in \cite{shuai-thesis} both reduces the number of particles and increases the estimation accuracy. Li et al. developed a particle filter that samples the contact state space through a contact graph and updates continuous state space with Kalman filters \cite{shuai-icra-2015}. We consider the work in this paper to be a continuous work on their work with the constraints, which are derived from the contact states, enforced in the Kalman filter update step.
\section{Derivation of the Contact-based RBPF}
In a multi-body dynamic system, the motions of the bodies can be formulated by the Newton-Euler equations and a set of complementary constraints, which are introduced by the intermittent contacts between the bodies. The equations are usually converted to their discretized form in order to be solved in time-stepping schemes. Additionally, if we further linearize the friction cone by approximating it with polygons/polyhedrons, the multi-body dynamic system can then be modeled as a Linear Complementarity Problem (LCP) model. A LCP model for a general multi-body dynamic system can be formulated as follows:
{\allowdisplaybreaks
\begin{align}
\label{LCP-NE-Equation}
\begin{bmatrix}
		0\\
		0\\
		(\rho_n)_{t+1}\\
		(\rho_f)_{t+1}\\
		s_{t+1}
	\end{bmatrix} &=
	\begin{bmatrix}
	-M & G_b &G_n &G_f &0\\
	G_b^T & 0 & 0& 0& 0\\
	G_n^T & 0 & 0 & 0 & 0\\
	G_f^T &0 &  0&  0 &E \\
	0 & 0 & U & -E^T & 0
	\end{bmatrix}
	\begin{bmatrix}
	v_{t+1}\\
	(p_b)_{t+1}\\
	(p_n)_{t+1}\\
	(p_f)_{t+1}\\
	\sigma_{t+1}
	\end{bmatrix} \\ \nonumber
	&+
	\begin{bmatrix}
	Mv_t + p_{app} + p_{vp}\\
	\frac{\Phi_t}{h} + \frac{\partial\Phi_t}{\partial t}\\
	\frac{(\Psi_n)_{t}}{h} + \frac{\partial(\Psi_n)_t}{\partial t}\\
	\frac{\partial(\Psi_f)_t}{\partial t}\\
	0
	\end{bmatrix} 	\\
	q_{t+1} &= q_t + G(\dot q)v_{t+1}h\\
	\label{LCP-Comp-Equation}
	& 0 \leq \begin{bmatrix} (\rho_n)_{t+1}\\ (\rho_f)_{t+1}\\s_{t+1}\end{bmatrix} \perp \begin{bmatrix} (p_n)_{t+1}\\ (p_f)_{t+1}\\ \sigma_{t+1}\end{bmatrix} \geq 0
\end{align}
}
where $M$ is the inertia matrix of all the objects, $v_{(\cdot)}$ is the vector of the velocities, $G_{(\cdot)}$ are the Jacobian matrices for the bilateral, contact normal and contact frictional constraints, $U$ is the friction coefficient matrix, $E$ is a selection matrix, $\Psi_{(\cdot)}$ and $\Phi_{(\cdot)}$ are gap distances for the unilateral and bilateral constraints, respectively, $p_{(\cdot)}$ is the impulse calculated as $f_{(\cdot)}h$, $\rho_{(\cdot)} = \Psi_{(\cdot)}/h$ and the subscripts represent the time steps. We evaluate all the Jacobian matrices $G_{(\cdot)}$ at time step $t$.

A solution to a LCP model consists of positive numbers and zeros assigned to vectors $\begin{bmatrix}(p_n)_{t+1} & (p_f)_{t+1} & \sigma_{t+1}\end{bmatrix}^T$ and $\begin{bmatrix} (\rho_n)_{t+1} & (\rho_f)_{t+1} & s_{t+1}\end{bmatrix}^T$, and to satisfy the complementarity conditions, wherever a element in vector $\begin{bmatrix}(p_n)_{t+1} & (p_f)_{t+1}& \sigma_{t+1}\end{bmatrix}^T$ is positive, its counter part element in vector $\begin{bmatrix} (\rho_n)_{t+1} & (\rho_f)_{t+1} & s_{t+1}\end{bmatrix}^T$ is zero and vice versa. The physical interpretation of this solution is that whenever there is a gap ($\rho_{(\cdot)} > 0$) between two contact points, the corresponding impulses ($p_{(\cdot)}$) are zero, and when there are impulses ($p_{(\cdot)}> 0$) between two contacts, the contact is formed and thus the gap ($\rho_{(\cdot)}$) is zero.

A contact state for a multi-body dynamic system is a complete list of all the contacts formed in the system. Therefore, given a contact state, the signs of the elements in vectors $\begin{bmatrix}(p_n)_{t+1} & (p_f)_{t+1} & \sigma_{t+1}\end{bmatrix}^T$ and $\begin{bmatrix} (\rho_n)_{t+1} & (\rho_f)_{t+1} & s_{t+1}\end{bmatrix}^T$ will be deterministic. As a result, the complementarity conditions will be converted into a set of equality constraints and inequality constraints\footnote{Since only $(p_n)_{t+1}$, $(p_f)_{t+1}$, $\sigma_{t+1}$ get involve with updating the actual state ($v_t$) in the Newton-Euler equations, we focus on expressing the constraints in terms of these variables.} with known a contact state. These equality (equations(\ref{eq:equality-constraints1}), (\ref{eq:equality-constraints2})) and inequality (equations(\ref{eq:inequality-constraints1}), (\ref{eq:inequality-constraints2})) constraints are shown below: 
{\allowdisplaybreaks
\begin{align}
\label{eq:equality-constraints1}
\begin{bmatrix}
\left((\rho_n)_{t+1}\right)_{\alpha}\\
\left((\rho_f)_{t+1}\right)_{\alpha}\\
\left(s_{t+1}\right)_{\alpha}
\end{bmatrix} = \begin{bmatrix}
0\\0\\0
\end{bmatrix} &= \begin{bmatrix}
(G_n^T)_{\alpha\cdot}v_{t+1}\\
(G_f^T)_{\alpha\cdot}v_{t+1} + E_{\alpha\alpha}\left(\sigma_{t+1}\right)_{\alpha}\\
U_{\alpha\alpha}\left((p_n)_{t+1}\right)_{\alpha} - E_{\alpha\alpha}\left((p_f)_{t+1}\right)_{\alpha}
\end{bmatrix} \\ \nonumber
&+ \begin{bmatrix}
\left(\frac{(\Psi_n)_{t}}{h}\right)_{\alpha}\\0\\0
\end{bmatrix}\\
\label{eq:equality-constraints2}
\begin{bmatrix}
0\\0\\0
\end{bmatrix} &= \begin{bmatrix}
\left((p_n)_{t+1}\right)_{\beta}\\
\left((p_f)_{t+1}\right)_{\beta}\\
\left(\sigma_{t+1}\right)_{\beta}
\end{bmatrix}\\
\label{eq:inequality-constraints1}
0 \leq \begin{bmatrix}
\left((\rho_n)_{t+1}\right)_{\beta}\\
\left((\rho_f)_{t+1}\right)_{\beta}\\
\left(s_{t+1}\right)_{\beta}
\end{bmatrix} &= \begin{bmatrix}
(G_n^T)_{\beta\cdot}v_{t+1}\\
(G_f^T)_{\beta\cdot}v_{t+1} + E_{\beta\alpha}\left(\sigma_{t+1}\right)_{\alpha}\\
U_{\beta\alpha}\left((p_n)_{t+1}\right)_{\alpha} - E_{\beta\alpha}\left((p_f)_{t+1}\right)_{\alpha}
\end{bmatrix} \\ \nonumber
&+ \begin{bmatrix}
\left(\frac{(\Psi_n)_{t}}{h}\right)_{\beta}\\0\\0
\end{bmatrix}\\
\label{eq:inequality-constraints2}
\begin{bmatrix}
\left((p_n)_{t+1}\right)_{\alpha}\\
\left((p_f)_{t+1}\right)_{\alpha}\\
\left(\sigma_{t+1}\right)_{\alpha}
\end{bmatrix} &\geq 0
\end{align}
}

where the subscripts ($\alpha, \beta, \cdot$) represent the subsets of indices in the number of rows and columns for the matrices (e.g., $G_n^T$) and vectors (e.g., $(p_n)_{t+1}$). Subset $\alpha$ represents the constraints that are enforced by the contact state. Subset $\beta$ represents the constraints that are excluded by the contact state. Subset $\cdot$ includes all the indices.

We combine the equality constraints (equations(\ref{eq:equality-constraints1}) and (\ref{eq:equality-constraints2})) with the Newton-Euler equation in the first row of equation (\ref{LCP-NE-Equation}). With some linear algebra manipulations, we are able to derive a linear transition model for the continuous state\footnote{The continuous state term is used to differentiate itself from the discrete contact state. Here ``continuous" refers to the state space being continuous.} (i.e., $v_t$) of the bodies, which is shown below: 
\begin{align}
v_{t+1} &= Av_t + Bu_t\\
\label{eq:Linear-A}
A &= M^{-1}(M - HK^{-1}H^T)\\
\label{eq:Linear-B}
B &= M^{-1}\begin{bmatrix} I-HK^{-1}F&-H\end{bmatrix}\\
\label{eq:linear-final}
H &= \begin{bmatrix}(G_n)_{\cdot\alpha} &(G_f)_{\cdot\alpha} &0\end{bmatrix}\\
F &= \begin{bmatrix} (\hat{G}_n)_{\alpha\cdot}\\(\hat{G}_f)_{\alpha\cdot}\\0\end{bmatrix}\\
\label{eq:K-matrix}
K &= \begin{bmatrix}(\hat{G}_nG_n)_{\alpha\alpha} &(\hat{G}_nG_f)_{\alpha\alpha} &0\\(\hat{G}_fG_n)_{\alpha\alpha} &(\hat{G}_fG_f)_{\alpha\alpha} &E_{\alpha\alpha}\\U_{\alpha\alpha} & -(E^T)_{\alpha\alpha} &0\end{bmatrix}\\
u_t &= \begin{bmatrix}p_{app}&(\frac{(\Psi_n)_t}{h})_{\alpha\cdot}&0&0\end{bmatrix}^T\\
\end{align}
where $\hat{G}_n = G_n^TM^{-1}$ and $\hat{G}_f = G_f^TM^{-1}$. With the linear transition model above, we will  be able to apply a Kalman filter to update the state $v_t$. The pose can then be updated using the updated $v_t$.

In summary, for a multi-body dynamic system, with a known contact state, a set of equality and inequality constraints can be derived from its LCP model formulation. Based on the equality constraints, we are able to derive a linear transition model for the continuous state ($v_t$) of the bodies in the system, which allows for Kalman filters to be used to update $v_t$. Therefore, a contact-based RBPF samples the contact state using particles and updates the continuous state with Kalman filters\footnote{Each particle runs a Kalman filter in parallel to other particles}.  Algorithm \ref{alg:rbpf-pseudo-code} shows the pseudo code of the contact-based RBPF, where for time step $t+1$, $c_{t+1}$ is the discrete contact state, $\hat{x}_{t+1}$ is mean of the continuous state (both pose and velocity), $P_{t+1}$ is the covariance matrix of the continuous state, $u_{t+1}$ is the input to the dynamic system, $z_{t+1}$ is the measurement, such as the measured pose of an object, $w_{t+1}$ is vector of weights of all particles, the superscripts indicate the ids of the particles the variables correspond to,  $Sample\_Contact\_State(\hat{x}_{t}^{[i]}$, $P_{t}^{[i]}$, $c_{t}^{[i]})$ samples the contact state for time step $t+1$, $Kalman\_Update(\hat{x}_{t}^{[i]}, P_t^{[i]}, c_{t + 1}^{[i]}, u_{t+1}, z_{t+1})$ updates the continuous states with a Kalman filter update step and $Weight\_Update(\hat{x}_{t+1}^{[i]}, P_{t+1}^{[i]}, z_{t+1}, w_{t}^{[i]})$ updates the weights of the particles according to the measurements $z_{t+1}$. 
\begin{breakablealgorithm}
\caption{Contact-Based RBPF}
\label{alg:rbpf-pseudo-code} 
\begin{algorithmic}
	\Function {Filter}{$\hat{x}_{t}$, $P_{t}$, $c_{t}$, $u_{t+1}$, $z_{t+1}$, $w_{t}$}
		\For{$i = 1 \to N$}
			\State $c_{t+1}^{[i]}$ = \Call{Sample\_Contact\_State}{$\hat{x}_{t}^{[i]}$, $P_{t}^{[i]}$, $c_{t}^{[i]}$}
			\State $\hat{x}_{t + 1}^{[i]}$, $P_{t + 1}^{[i]}$ = \Call{Kalman\_Update}{$\hat{x}_{t}^{[i]}$, $P_t^{[i]}$, $c_{t + 1}^{[i]}$, $u_{t+1}$, $z_{t+1}$}
			\State $w_{t+1}^{[i]}$ = \Call{Weight\_Update}{$\hat{x}_{t+1}^{[i]}$, $P_{t+1}^{[i]}$, $z_{t+1}$, $w_{t}^{[i]}$}
		\EndFor
		\If{resample condition satisfied}
			\For{$i = 1 \to N$}
				\State Draw j with probability $\propto w_{t+1}^{[j]}$
				\State $\hat{x}_{t+1}^{'[i]}$ =  $\hat{x}_{t + 1}^{[j]}$, $c_{t+1}^{'[i]} = c_{t+1}^{[j]}$, $P_{t+1}^{'[i]} = P_{t+1}^{[j]}$, $w_{t+1}^{'[i]}$ = $\frac{1}{N}$
			\EndFor
		\Else
			\State $\hat{x}'_{t+1} = \hat{x}_{t+1}$, $c'_{t+1} = c_{t+1}$, $P'_{t+1} = P_{t+1}$, $w'_{t+1} = w_{t+1}$
		\EndIf
		\State \Return $\hat{x}'_{t+1}$, $P'_{t+1}$, $c'_{t+1}$, $w'_{t+1}$
	\EndFunction
\end{algorithmic}
\end{breakablealgorithm} 
\section{Derivation of the Constrained Contact-based RBPF}
As one may notice, although the equality constraints (equation (\ref{eq:equality-constraints1}) and equation (\ref{eq:equality-constraints2})) are used to derive the linear transition model, these constraints are not enforced in the Kalman filters during the continuous state update. Additionally, the inequality constraints (equation (\ref{eq:inequality-constraints1}) and equation (\ref{eq:inequality-constraints2})) are not enforced either. This unconstrained Kalman filter update has two problems. First, the equality and inequality constraints define a physical feasible sub-space in the continuous state space imposed by a contact state. Ignoring these constraints leads to assigning positive probability mass to invalid states. As a result, both the efficiency of the particle sampling and the accuracy of the distribution of continuous state will be downgraded. Second, one of the main reasons of instabilities in multi-body dynamic simulations is the inaccuracy of the collision detection algorithms. This inaccuracy can cause unfeasible object states, e.g., overlaps between two rigid bodies, which can dramatically damage the stability of the simulations. For example, the overlaps between two rigid bodies can introduce significant separation forces between the overlapping bodies, which thus generate large separation velocities for both bodies. Since our linear transition model derived above is based on the LCP model, unfeasible object states also introduce instabilities to our filter. The most common instability caused by unfeasible object states is the ``explosive" separation velocities due to rigid bodies overlapping. In consequence, the particles that suffer from such instabilities diverge significantly from the measurements and thus it can lead to particle starvation.

To tackle the two problems mentioned above, we propose the constrained contact-based RBPF. Similar to the contact-based RBPF, the constrained contact-based RBPF also samples the contact states with particles. However, the constrained contact-based RBPF updates the continuous state through Kalman filters with equality and inequality constraints enforced from equations (\ref{eq:equality-constraints1}, \ref{eq:equality-constraints2}, \ref{eq:inequality-constraints1}, \ref{eq:inequality-constraints2}). 

\subsection{Equality and Inequality Constraints}
In equation (\ref{eq:equality-constraints1}), we discover that there are two types of equality constraints: constraints that are related to state $x_{t+1}$ and constraints that are enforced on state $x_t$. We term the  set of equality constraints that are related to $x_{t+1}$ as  $S_{t+1}$ and the set of the constraints that are related to $x_t$ as $\tilde{S}_{t+1}$. From equation (\ref{eq:equality-constraints1}), we can derive $S_{t+1}$ as a linear equation $A_{t+1}^Px_{t+1} = b_{t+1}$ with matrix $A_{t+1}^P$ and vector $b_{t+1}^P$ defined as follows:
\begin{align}
\label{eq:equality-constraint-t+1-1}
A_{t+1}^P &= \begin{bmatrix}
(G_n^T)_{\alpha\cdot}\\
(G_f^T)_{\alpha\cdot}
\end{bmatrix}\\
\label{eq:equality-constraint-t+1-2}
b_{t+1}^P &= \begin{bmatrix}
-\left(\frac{(\Psi_n)_{t}}{h}\right)_{\alpha}\\
-E_{\alpha\alpha}\left(\sigma_{t+1}(x_t)\right)_{\alpha}
\end{bmatrix}
\end{align}
In the equations above, although $\sigma_{t+1}$ is an unknown variable, it can be expressed in terms of the state $x_t$ and thus we express it as $\sigma_{t+1}(x_t)$. 

Notice that among all the equality constraints in equations (\ref{eq:equality-constraints1}, \ref{eq:equality-constraints2}), only the first two rows of equation (\ref{eq:equality-constraints1}) are related to $x_{t+1}$ and are used to derive the equality constraint $S_{t+1}$. On the other hand, \cite{shuai-thesis} shows that the impulse vectors $(p_{(\cdot)})_{t+1}$ can be expressed in terms of the continuous state vector $x_t$.  Since the rest of the equality constraints in equations (\ref{eq:equality-constraints1}, \ref{eq:equality-constraints2}) are related to $(p_{(\cdot)})_{t+1}$, these equality constraints are enforced on the state vector $x_t$ and these equality constraints ($\tilde{S}_{t+1}$) are shown as below:
\begin{align}
\label{eq:equality-constraint-t-1}
U_{\alpha\alpha}\left((p_n)_{t+1}(x_t)\right)_{\alpha} - E_{\alpha\alpha}\left((p_f)_{t+1}(x_t)\right)_{\alpha} &= 0\\
\label{eq:equality-constraint-t-2}
\left((p_n)_{t+1}(x_t)\right)_{\beta} &= 0\\
\label{eq:equality-constraint-t-3}
\left((p_f)_{t+1}(x_t)\right)_{\beta} &= 0\\
\label{eq:equality-constraint-t-4}
\left(\sigma_{t+1}(x_t)\right)_{\beta} &= 0
\end{align}
$\tilde{S}_{t+1}$ can be interpreted as the following: with the new discoveries, i.e., the estimated contact state, at time step $t+1$, the continuous state estimation from time step $t$ will need to be modified in order to make the state transition between the two time steps feasible. As described in \cite{shuai-thesis}, $(p_f)_{t+1}(x_t)$, $(p_n)_{t+1}(x_t)$ and $\sigma_{t+1}(x_t)$ can all be expressed linearly in terms of $x_t$.  Therefore, equations (\ref{eq:equality-constraint-t-1}, \ref{eq:equality-constraint-t-2}, \ref{eq:equality-constraint-t-3}, \ref{eq:equality-constraint-t-4}) can be expressed as a linear constraint:
\begin{align}
A_{t\mid t+1}^P x_t = b_{t\mid t+1}^P 
\end{align}
where $A_{t\mid t+1}^P$ and $b_{t\mid t+1}^P$ are derived according to the method in \cite{shuai-thesis} and the subscript $t\mid t+1$ indicates that the constraints are for state $x_t$ with the information from time step $t+1$.

Similar to the equality constraints, the inequality constraints are also divided into two groups $R_{t+1}$ and $\tilde{R}_{t+1}$, where $R_{t+1}$  corresponds to the set of inequality constraints (\ref{eq:inequality-constraint-t+1-1}, \ref{eq:inequality-constraint-t+1-2}, \ref{eq:inequality-constraint-t+1-3}) for $x_{t+1}$ and $\tilde{R}_{t+1}$ is the set of inequality constraints (\ref{eq:inequality-constraint-t-1}, \ref{eq:inequality-constraint-t-2}, \ref{eq:inequality-constraint-t-3}, \ref{eq:inequality-constraint-t-4}) that are related to $x_{t}$. Again, following \cite{shuai-thesis}, $A_{t+1}^Q$, $b_{t+1}^Q$, $A_{t\mid t+1}^Q$ and $b_{t\mid t+1}^Q$ are derived for $R_{t+1}$ and $\tilde{R}_{t+1}$ as follows:
\begin{align}
\label{eq:inequality-constraint-t+1-1}
A_{t+1}^Qx_{t+1} &\ge b_{t+1}^Q\\
\label{eq:inequality-constraint-t+1-2}
A_{t+1}^Q &= \begin{bmatrix}
(G_n^T)_{\beta\cdot}\\
(G_f^T)_{\beta\cdot}
\end{bmatrix}\\
\label{eq:inequality-constraint-t+1-3}
b_{t+1}^Q &= \begin{bmatrix}
-\left(\frac{(\Psi_n)_{t}}{h}\right)_{\beta}\\
-E_{\beta\alpha}\left(\sigma_{t+1}(x_t)\right)_{\alpha}
\end{bmatrix}
\end{align}
\begin{align}
\label{eq:inequality-constraint-t-1}
U_{\beta\alpha}\left((p_n)_{t+1}(x_t)\right)_{\alpha} - E_{\beta\alpha}\left((p_f)_{t+1}(x_t)\right)_{\alpha} &\ge 0\\
\label{eq:inequality-constraint-t-2}
\left((p_n)_{t+1}(x_t)\right)_{\alpha} &\ge 0\\
\label{eq:inequality-constraint-t-3}
\left((p_f)_{t+1}(x_t)\right)_{\alpha} &\ge 0\\
\label{eq:inequality-constraint-t-4}
\left(\sigma_{t+1}(x_t)\right)_{\alpha} &\ge 0\\
A_{t\mid t+1}^Q x_t \ge  b_{t\mid t+1}^Q
\end{align}

\subsection{Update the Continuous State with Equality and Inequality Constraints}
For time step $t+1$, we have the mean of the Kalman filter updated unconstrained continuous state  as $\hat{x}_{t+1}$ and the covariance matrix as $P_{t+1}$. Similarly, for the constrained continuous state, we define its mean as $\hat{x}^C_{t+1}$ and its covariance matrix as $P_{t+1}^C$. With the equality constraints in $S_{t+1}$ (defined in equations (\ref{eq:equality-constraint-t+1-1}, \ref{eq:equality-constraint-t+1-2})), the constrained continuous state vector mean $\hat{x}^C_{t+1}$ and the covariance matrix $P_{t+1}^C$ can be calculated following the derivation in \cite{Gupta2008}, where $\hat{x}^C_{t+1}$ can be computed as the solution to the following quadratic programming problem:
\begin{align}
\hat{x}^C_{t+1} &= \argmin_{x \in R^n}(x - \hat{x}_{t+1})^TW(x - \hat{x}_{t+1})\\
s.t.\ &A_{t+1}^Px = b_{t+1}^P
\end{align} 
where $W$ is a positive definite symmetric weighting matrix. The Karush-Kuhn-Tucker (KKT) condition of the above optimization problem is as follows:
\begin{align}
\centering
\label{eq:KKT-condition}
\begin{bmatrix}
2W&A_{t+1}^P\\
A_{t+1}^P&0
\end{bmatrix}
\begin{bmatrix}
X\\
\lambda
\end{bmatrix}=\begin{bmatrix}
0\\
-(A_{t+1}^P\hat{x}_{t+1} - b_{t+1}^P)
\end{bmatrix}
\end{align}
where $X = x  - \hat{x}_{t+1}$ and $\lambda$ is the Lagrangian multiplier. With some algebraic manipulation to the KKT condition above, the solution to the above minimization problem is found as the following:
\begin{align}
\label{eq:constrained-mean}
\hat{x}^C_{t+1} =& \hat{x}_{t+1} - W^{-1}(A_{t+1}^P)^T\left(A_{t+1}^PW^{-1}(A_{t+1}^P)^T\right)^{-1}...\\
&...(A_{t+1}^P\hat{x}_{t+1} - b_{t+1}^P)
\end{align}
In the solution above, the unconstrained mean vector $\hat{x}_{t+1}$ is projected onto the constraint plane $A_{t+1}^Px = b_{t+1}^P$ with the projection matrix $(I - \gamma_{t+1}A_{t+1}^P)$, where matrix $\gamma_{t+1}$ is defined as below:
\begin{align}
\gamma_{t+1} =  W^{-1}(A_{t+1}^P)^T\left(A_{t+1}^PW^{-1}(A_{t+1}^P)^T\right)^{-1}
\end{align}
As a result, the solution to the mean of the constrained Kalman filter can be simplified as $\hat{x}^C_{t+1} = \hat{x}_{t+1} - \gamma_{t+1}(A_{t+1}^P\hat{x}_{t+1} - b_{t+1}^P)$ and the constrained covariance matrix $P_{t+1}^C$ can be derived as the following:
\begin{align}
\label{eq:constraint-equality-covariance}
P_{t+1}^C = (I - \gamma_{t+1}A_{t+1}^P)P_{t+1}
\end{align}

Also, we can formulate the quadratic programming problem for state $x_t$ with constraint set $\tilde{S}_{t+1}$, which includes matrix $A_{t\mid t+1}^P$ and vector $b_{t\mid t+1}^P$, and solve for $\hat{x}^C_{t\mid t+1}$ and  $P_{t\mid t+1}^C$.

Similarly, the inequality constraints $R_{t+1}$ and $\tilde{R}_{t+1}$ can also be used to formulate two quadratic programming problems. However, as pointed out in \cite{Gupta2008}, in the case of inequality constraints, while the constrained mean $\hat{x}^C_{t+1} $ can be computed using standard quadratic programming methods, such as interior point method, computing the constrained covariance matirx $P_{t+1}^C$ with inequality constraints becomes much more difficult. 

In our constrained contact-based RBPF, we formulate a quadratic programming problem with the equality ($S_{t+1}$) and the inequality ($R_{t+1}$) constraints\footnote{We use constraint sets $S_{t+1}$ and $R_{t+1}$ as an example. The quadratic programming problem for constraint sets $\tilde{S}_{t+1}$ and $\tilde{R}_{t+1}$ can be formulated similarly. } as shown in equation (\ref{eq:constrained-qp}). The input to this quadratic programming problem is updated state from the Kalman filter and the solution to this quadratic programming problem will be the updated mean of the state with the equality and inequality constraints  incorporated. Similar to \cite{Gupta2008}, the covariance matrix is updated only with the equality constraints, which is shown in equation (\ref{eq:constraint-equality-covariance}).
\begin{align}
\label{eq:constrained-qp}
\hat{x}^C_{t+1} &= \argmin_{x \in R^n}(x - \hat{x}_{t+1})^TW(x - \hat{x}_{t+1})\\
s.t.\ &A_{t+1}^Px = b_{t+1}^P\\
	  &A_{t+1}^Qx \ge b_{t+1}^Q
\end{align}

Additionally, since we have constraints for both $x_{t+1}$ ($S_{t+1}$ and $R_{t+1}$) and $x_t$ ($\tilde{S}_{t+1}$ and $\tilde{R}_{t+1}$), in order to update the state $x_{t+1}$,  we propose to first update the state vector $x_t$ with constraints $\tilde{S}_{t+1}$ and $\tilde{R}_{t+1}$ before transitioning it to $x_{t+1}$ with a Kalman filter update. Then after the Kalman filter update, $x_{t+1}$ is updated with constraints $S_{t+1}$ and $R_{t+1}$. Therefore, based on the algorithm of contact-based RBPF in Algorithm \ref{alg:rbpf-pseudo-code}, we summarize our constrained contact-based RBPF as below:
\begin{breakablealgorithm}
\caption{Constrained Contact-Based RBPF}
\label{alg:c-rbpf-pseudo-code} 
\begin{algorithmic}
	\Function {Filter}{$\hat{x}_{t}$, $P_{t}$, $c_{t}$, $u_{t+1}$, $z_{t+1}$, $w_{t}$}
		\For{$i = 1 \to N$}
			\State $c_{t+1}^{[i]}$ = \Call{Sample\_Contact\_State}{$\hat{x}_{t}^{[i]}$, $P_{t}^{[i]}$, $c_{t}^{[i]}$}
			\State $\hat{x}_{t\mid t+1}^{C[i]}$, $P_{t\mid t+1}^{C[i]}$ = \Call{Solve\_QP}{$\hat{x}_t^{[i]}$, $P_t^{[i]}$, $\tilde{S}_{t+1}^{[i]}$,  $\tilde{R}_{t+1}^{[i]}$}
			\State $\hat{x}_{t + 1}^{[i]}$, $P_{t+1}^{[i]}$ = \Call{Kalman\_Update}{$\hat{x}_{t\mid t+1}^{C[i]}$, $P_{t\mid t+1}^{C[i]}$, $c_{t + 1}^{[i]}$, $u_{t+1}$, $z_{t+1}$}
			\State $\hat{x}_{t+1}^{C[i]}$, $P_{t+1}^{C[i]}$ = \Call{Solve\_QP}{$\hat{x}_{t+1}^{[i]}$, $P_{t+1}^{[i]}$, $S_{t+1}^{[i]}$, $R_{t+1}^{[i]}$}
			\State $w_{t+1}^{[i]}$ = \Call{Weight\_Update}{$\hat{x}_{t+1}^{C[i]}$,  $P_{t+1}^{C[i]}$, $z_{t+1}$, $w_{t}^{[i]}$}
		\EndFor
		\If{resample condition satisfied}
			\For{$i = 1 \to N$}
				\State Draw j with probability $\propto w_{t+1}^{[j]}$
				\State $\hat{x}_{t+1}^{'[i]}$ =  $\hat{x}_{t + 1}^{C[j]}$, $c_{t+1}^{'[i]} = c_{t+1}^{[j]}$, $P_{t+1}^{'[i]} = P_{t+1}^{C[j]}$, $w_{t+1}^{'[i]}$ = $\frac{1}{N}$
			\EndFor
		\Else
			\State $\hat{x}'_{t+1} = \hat{x}_{t+1}^C$, $c'_{t+1} = c_{t+1}$, $P'_{t+1} = P_{t+1}^C$, $w'_{t+1} = w_{t+1}$
		\EndIf
		\State \Return $\hat{x}'_{t+1}$, $P'_{t+1}$, $c'_{t+1}$, $w'_{t+1}$
	\EndFunction
\end{algorithmic}
\end{breakablealgorithm} 
where all the variables are defined in the same way with Algorithm \ref{alg:rbpf-pseudo-code} and $Solve\_QP(\hat{x}_{t+1}^{[i]}, P_{t+1}^{[i]}, S_{t+1}^{[i]}, R_{t+1}^{[i]})$ solves the quadratic programming problem defined in equation (\ref{eq:constrained-qp}). 

\section{Experiment Results and Discussions}
\begin{figure} [h!]
\centering
\begin{subfigure}{.8\linewidth}
  \centering
  \includegraphics[width=1\linewidth]{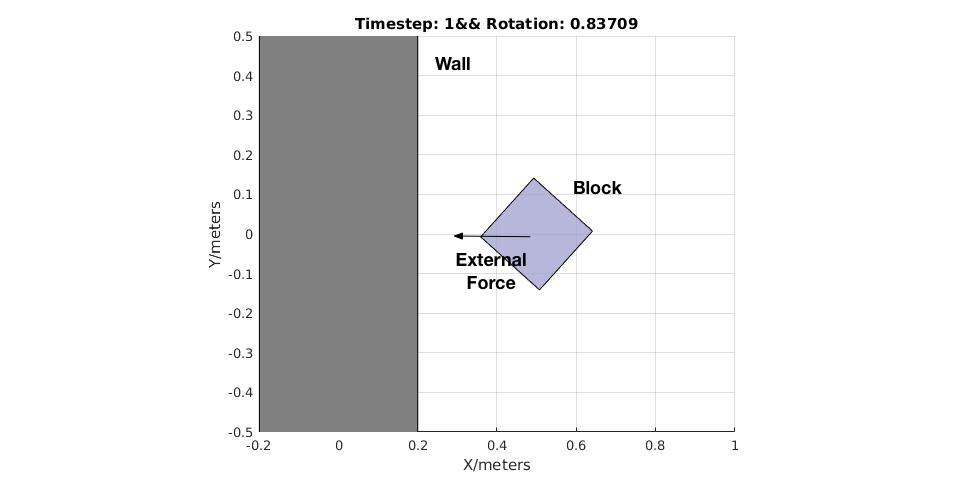}
  \caption{The simulation experiment where a block moves into contact with a wall.}
  \label{fig:wall-exp-setup}
\end{subfigure}
\hfill 
\begin{subfigure}{.8\linewidth}
  \centering
  \includegraphics[width=1\linewidth]{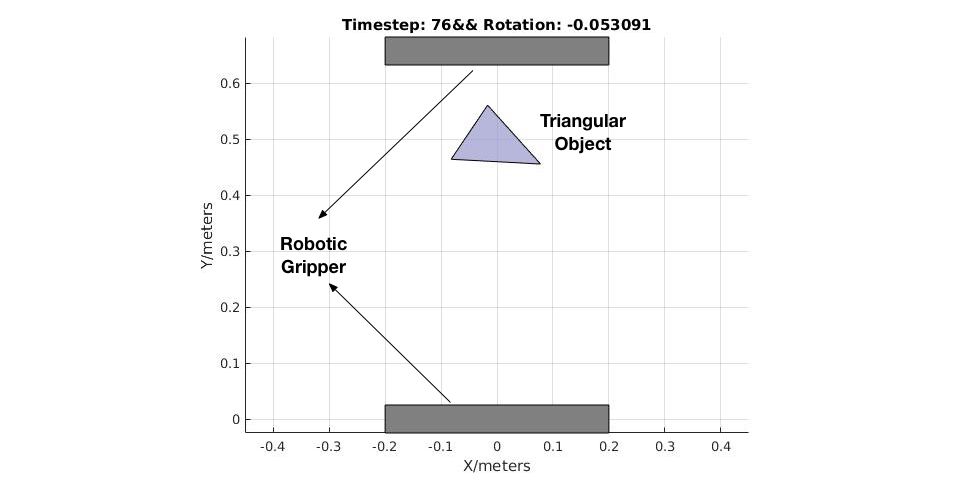}
 \caption{The simulation experiment where a robotic gripper grasps a triangular object.}
   \label{fig:gripper-exp-setup}
\end{subfigure}
\caption{The setup of the simulation experiments}
\end{figure}

Since our constrained contact-based RBPF is designed to improve the contact-based RBPF, we tested both filters in simulation experiments. The simulation experiments are performed using the RPI-Matlab simulator \cite{rpi-matlab-simulator} with the LCP model as its physics engine. Two set of simulation experiments were conducted: 
\begin{itemize}
\item A block, which is under an external force, moves into contact with a fixed wall (Figure \ref{fig:wall-exp-setup}).
\item A robotic gripper grasps a triangular object (Figure \ref{fig:gripper-exp-setup}).
\end{itemize}
\begin{figure}[h!]
\centering
\begin{subfigure}{.5\linewidth}
  \centering
  \includegraphics[width=.95\linewidth]{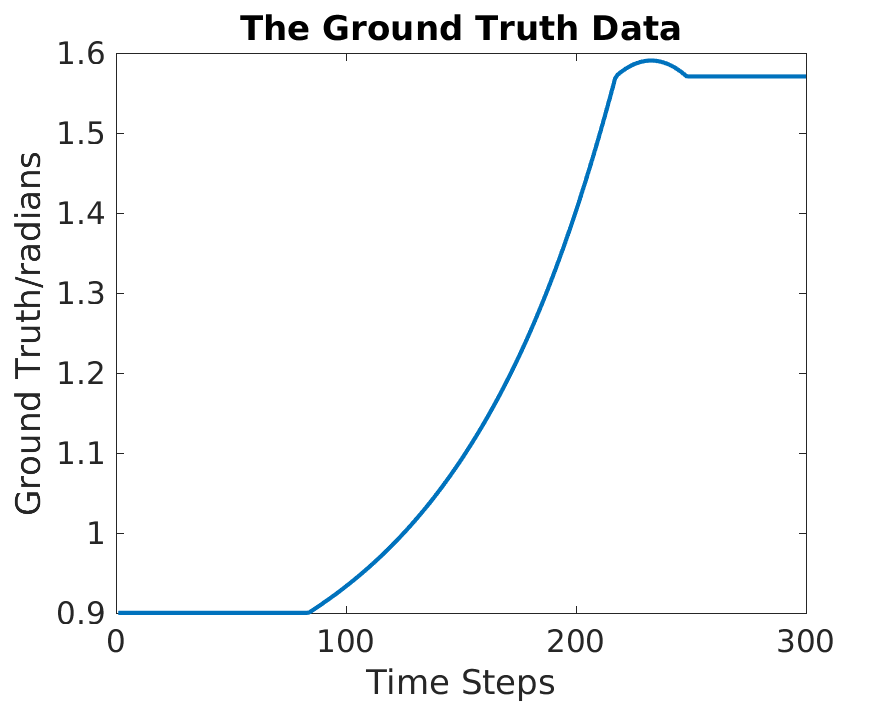}
  \caption{An example of the ground truth data.}
\end{subfigure}%
\begin{subfigure}{.5\linewidth}
  \centering
  \includegraphics[width=.95\linewidth]{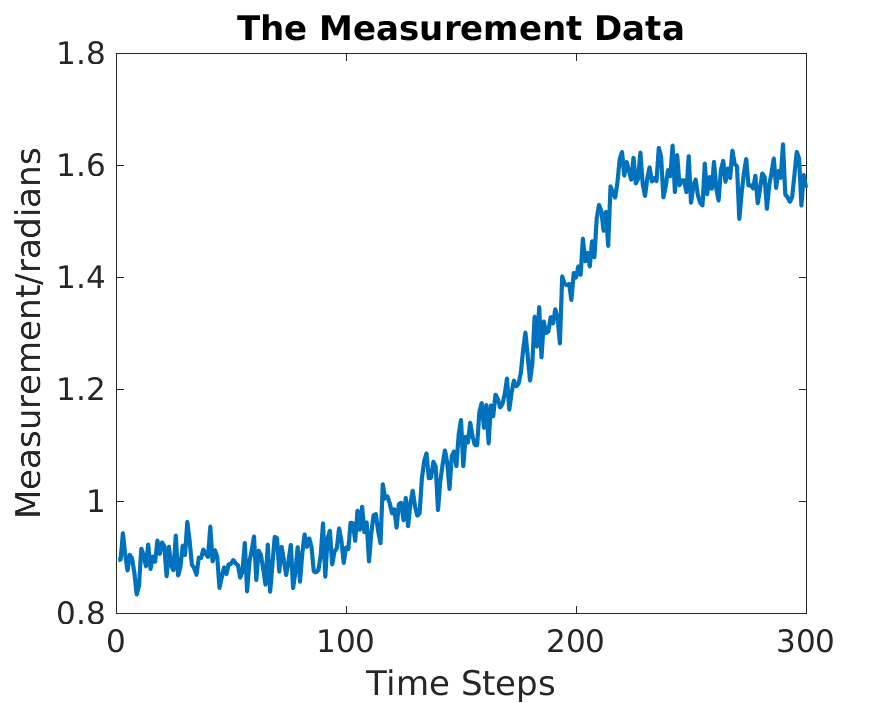}
 \caption{An example of the noisy measurement data.}
\end{subfigure}
\caption{A comparison between the ground truth data and its corresponding noisy measurements for rotations. }
 \label{fig:comparision-truth-measurement}
\end{figure}
In all simulation experiments, the ground truth data were acquired directly from the simulation engine and a white Gaussian noise (with 3cm variance on translation and 0.3 radians on rotation)  were added to the ground truth data to simulate the noisy measurements. An example of the level of noise in the measurements of the objects' rotations is shown in Figure \ref{fig:comparision-truth-measurement}, where a plot of the ground truth data is compared against its noisy measurements. In the experiments, only the poses of the block and the triangular object were measured and the velocity estimations (as will be seen in the next section) were inferred from the pose measurements.

\subsection{Results and Discussions}
\begin{figure*} [h!]
\centering
\begin{subfigure}{.33\linewidth}
  \centering
  \includegraphics[width=.95\linewidth]{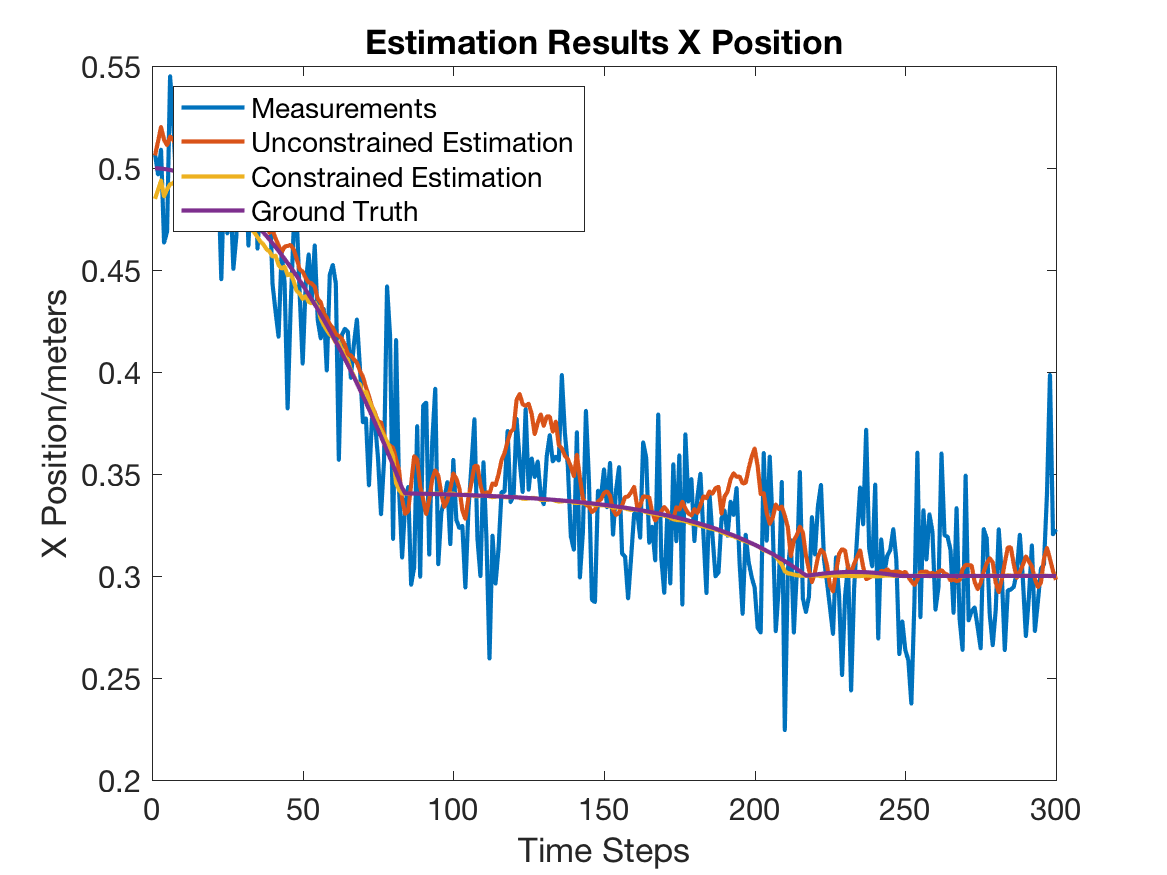}
  \caption{Estimation result for the x position of the block.}
  \label{fig:estimation-results-block-x}
\end{subfigure}%
\begin{subfigure}{.33\linewidth}
  \centering
  \includegraphics[width=.95\linewidth]{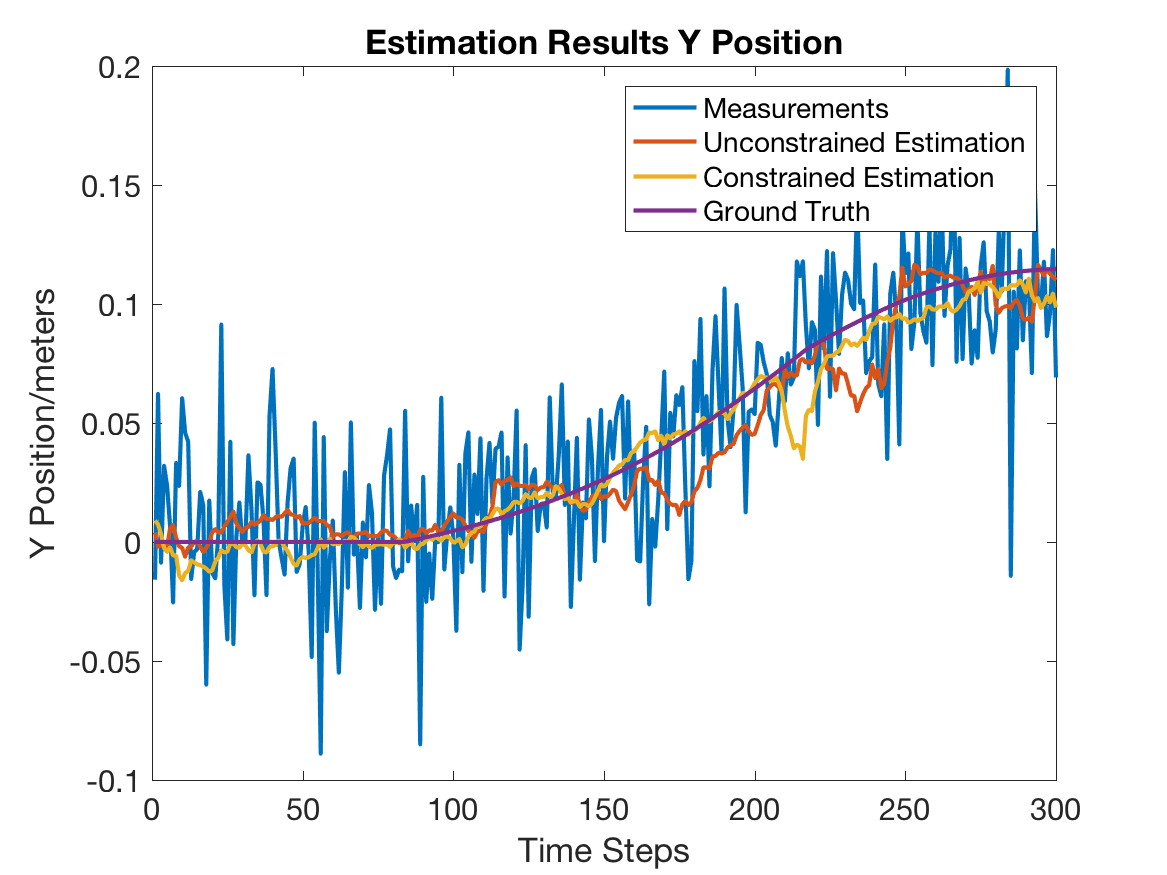}
  \caption{Estimation result for the y position of the block.}
  \label{fig:estimation-results-block-y}
\end{subfigure}%
\begin{subfigure}{.33\linewidth}
  \centering
  \includegraphics[width=.95\linewidth]{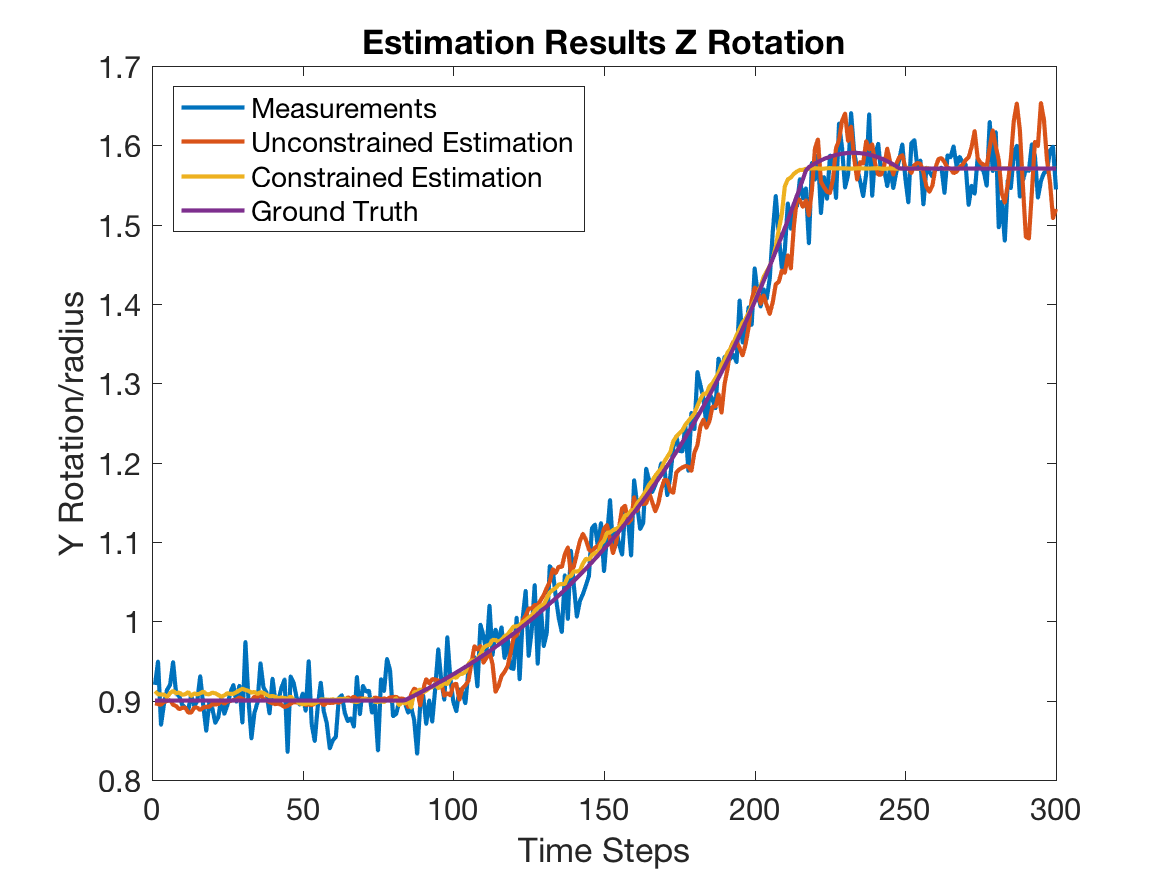}
  \caption{Estimation result for the z rotation of the block.}
  \label{fig:estimation-results-block-z}
\end{subfigure}
\hfill\begin{subfigure}{.33\linewidth}
  \centering
  \includegraphics[width=.95\linewidth]{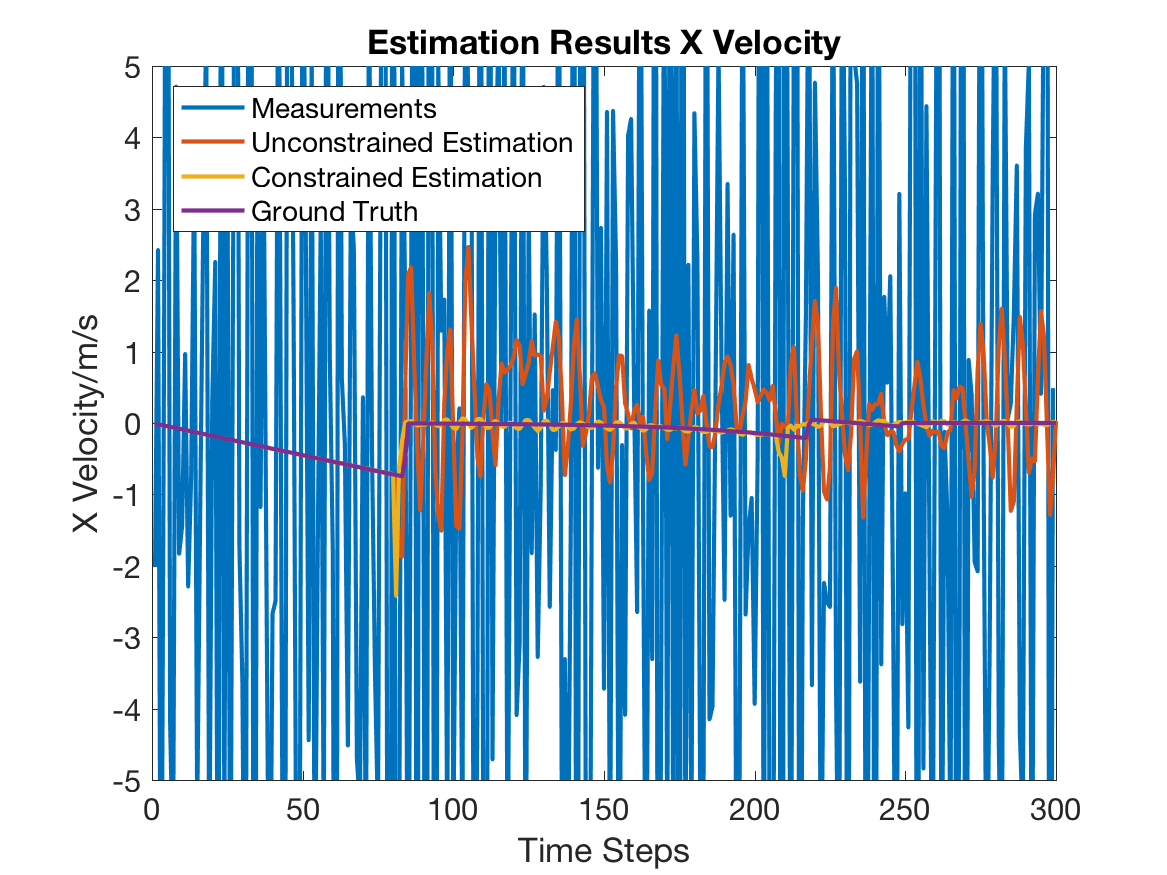}
  \caption{Estimation result for the x velocity of the block.}
  \label{fig:estimation-results-block-xv}
\end{subfigure}%
\begin{subfigure}{.33\linewidth}
  \centering
  \includegraphics[width=.95\linewidth]{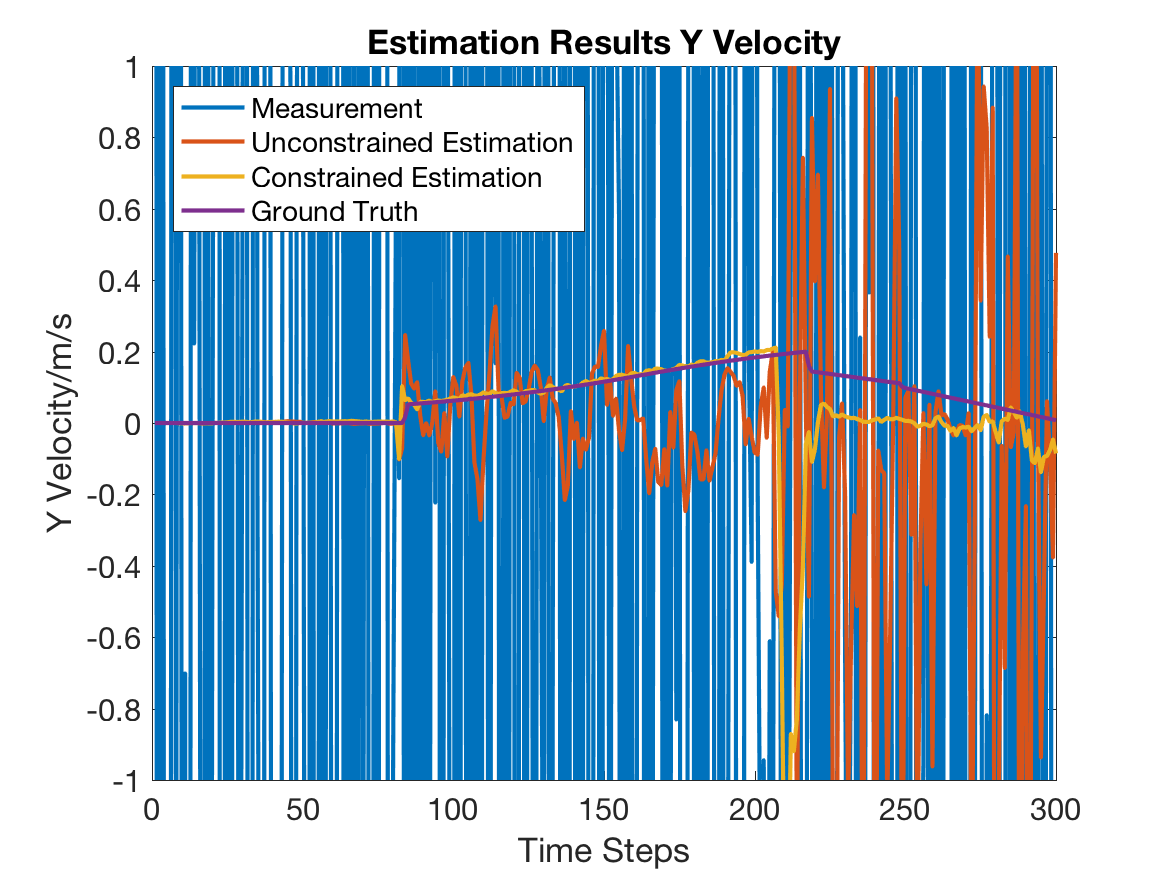}
  \caption{Estimation result for the y velocity of the block.}
  \label{fig:estimation-results-block-yv}
\end{subfigure}
\hfill\begin{subfigure}{.33\linewidth}
  \centering
  \includegraphics[width=.95\linewidth]{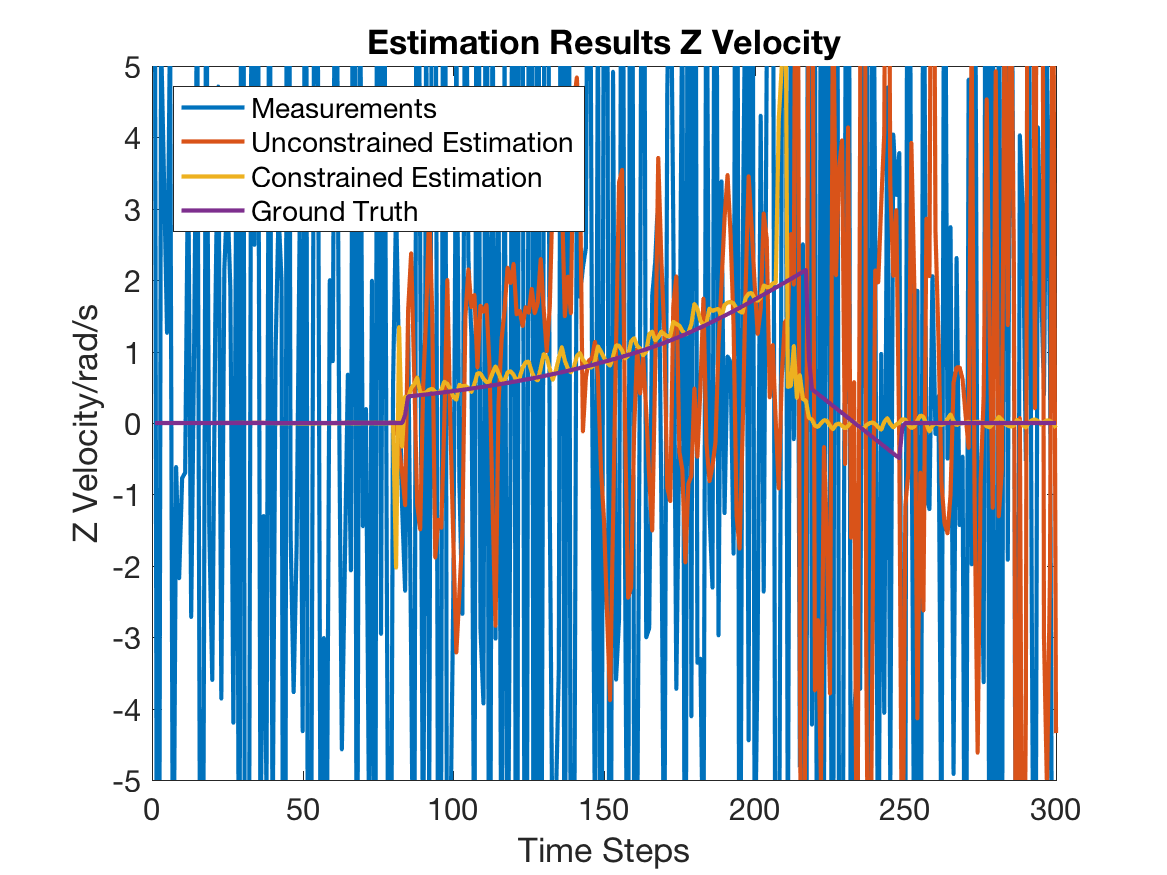}
  \caption{Estimation result for the z velocity of the block.}
  \label{fig:estimation-results-block-zv}
\end{subfigure} 
\caption{Estimation results for the first set of simulation experiments.}
\label{fig:estimation-results-block}
\end{figure*}

For the first set of simulation experiments, we ran both contact-based RBPF and constrained contact-based RBPF on the noisy measurements. The results are shown in Figure \ref{fig:estimation-results-block}, where the measured trajectories (blue), the estimation trajectories of both the contact-based RBPF (red) and the constrained contact-based RBPF (yellow) are overlaid with the ground truth trajectory (purple).

For the pose estimation results, Figure \ref{fig:estimation-results-block-x} and \ref{fig:estimation-results-block-z} show that the estimated trajectory of the constrained contact-based RBPF is smother and more accurate than that of the contact-based RBPF and Figure \ref{fig:estimation-results-block-y} shows that both filters perform similarly with constrained contact-based RBPF slightly better.  We find that when a state variable is constrained by certain constraints, enforcing those constraints in the constrained contact-based RBPF helps to improve the accuracy of estimation on the state variable. The block makes contact with the wall at about time step 80. The contact constraint together with the external force retain the block's translation along the x-axis and the rotation around the z-axis. Before time step 80 without the constraints, the performances of both filters are similar and the constrained contact-based RBPF outperforms the contact-based RBPF right after the contact formed. We also found that the covariance values of the constrained contact-based RBPF is much smaller than that of the contact-based RBPF in both translation and rotation estimation results. Additionally, since there are only frictional constraints, which are weaker compared with the contact constraints, on the translation along the y-axis, the constrained contact-based RBPF only slightly outperforms the contact-based RBPF. 

For the velocity estimation results, the measured velocity trajectories are induced by the position differences. Figure \ref{fig:estimation-results-block-xv}, \ref{fig:estimation-results-block-yv} and \ref{fig:estimation-results-block-zv} show that although both filters outperforms the induced velocity trajectories, the constrained contact-based RBPF significantly outperforms contact-based RBPF, where constrained contact-based RBPF can track the velocity trajectory fairly accurately while contact-based RBPF at its best (Figure \ref{fig:estimation-results-block-xv}) can only estimate the general trend of the velocity trajectory. We also notice that the constrained  contact-based RBPF generates spikes in its trajectory estimation during contact state transition (time steps 90 and 210). This is due to the uncertainties of the contact state estimation during the contact state transition period and we believe this can be alleviated by improving the contact state estimation.
\begin{figure*} [h!]
\centering
\begin{subfigure}{.33\textwidth}
  \centering
  \includegraphics[width=.95\linewidth]{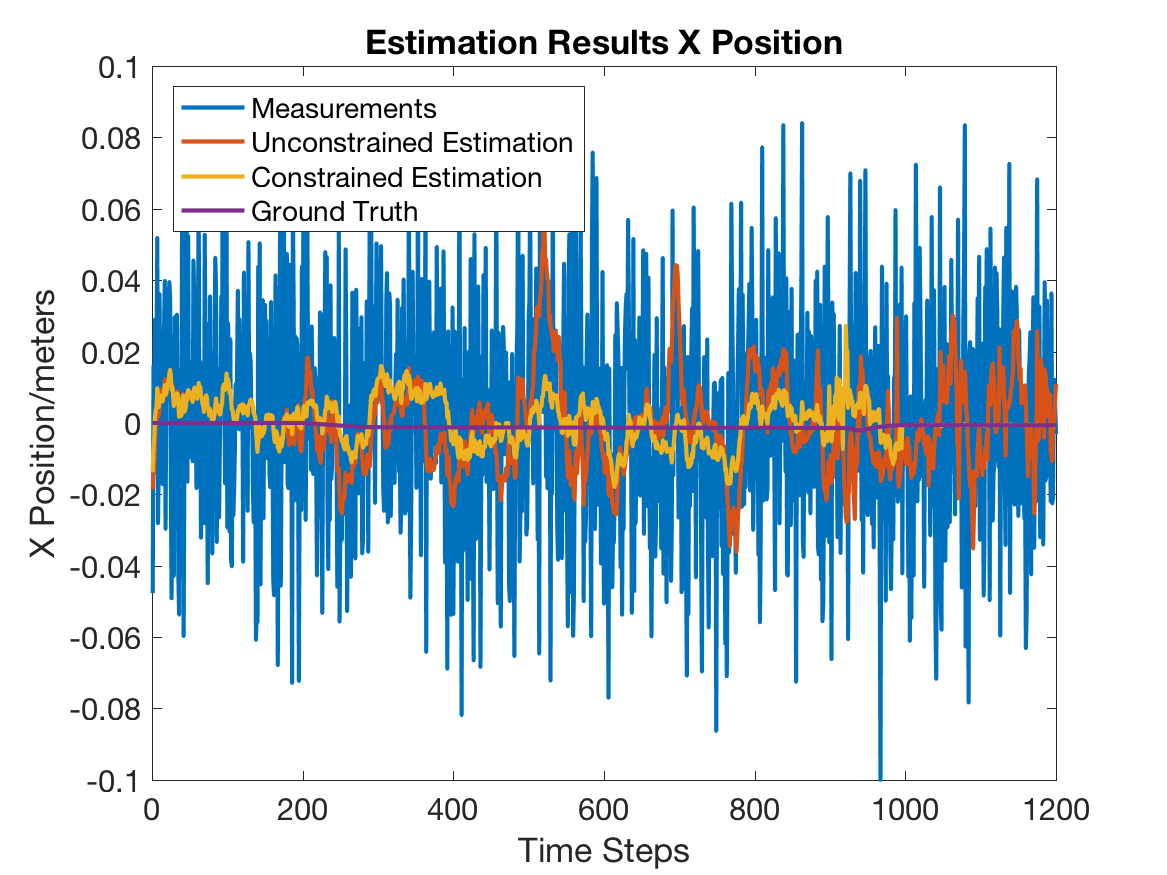}
  \caption{Estimation result for the x position of the object in the gripper.}
  \label{fig:estimation-results-gripper-x}
\end{subfigure}%
\begin{subfigure}{.33\textwidth}
  \centering
  \includegraphics[width=.95\linewidth]{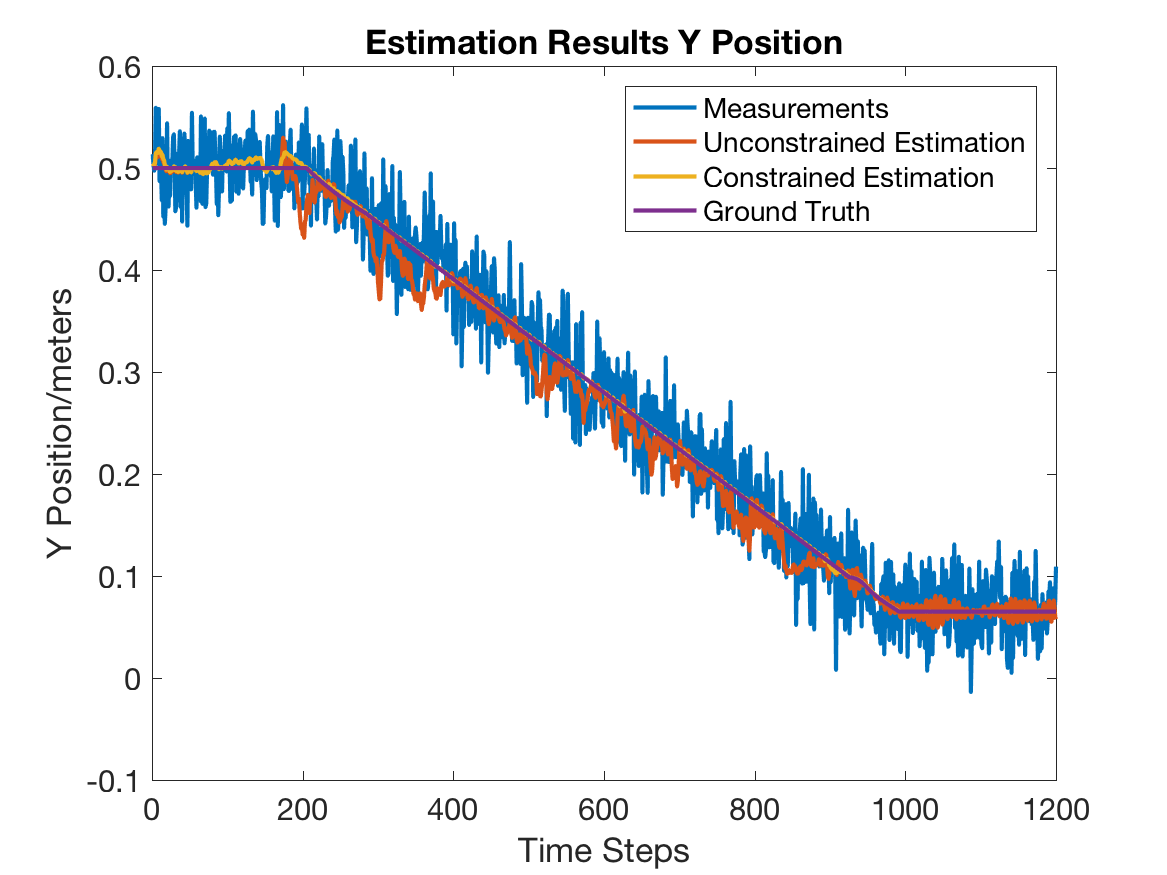}
  \caption{Estimation result for the y position of the object in the gripper.}
  \label{fig:estimation-results-gripper-y}
\end{subfigure}
\begin{subfigure}{.33\textwidth}
  \centering
  \includegraphics[width=.95\linewidth]{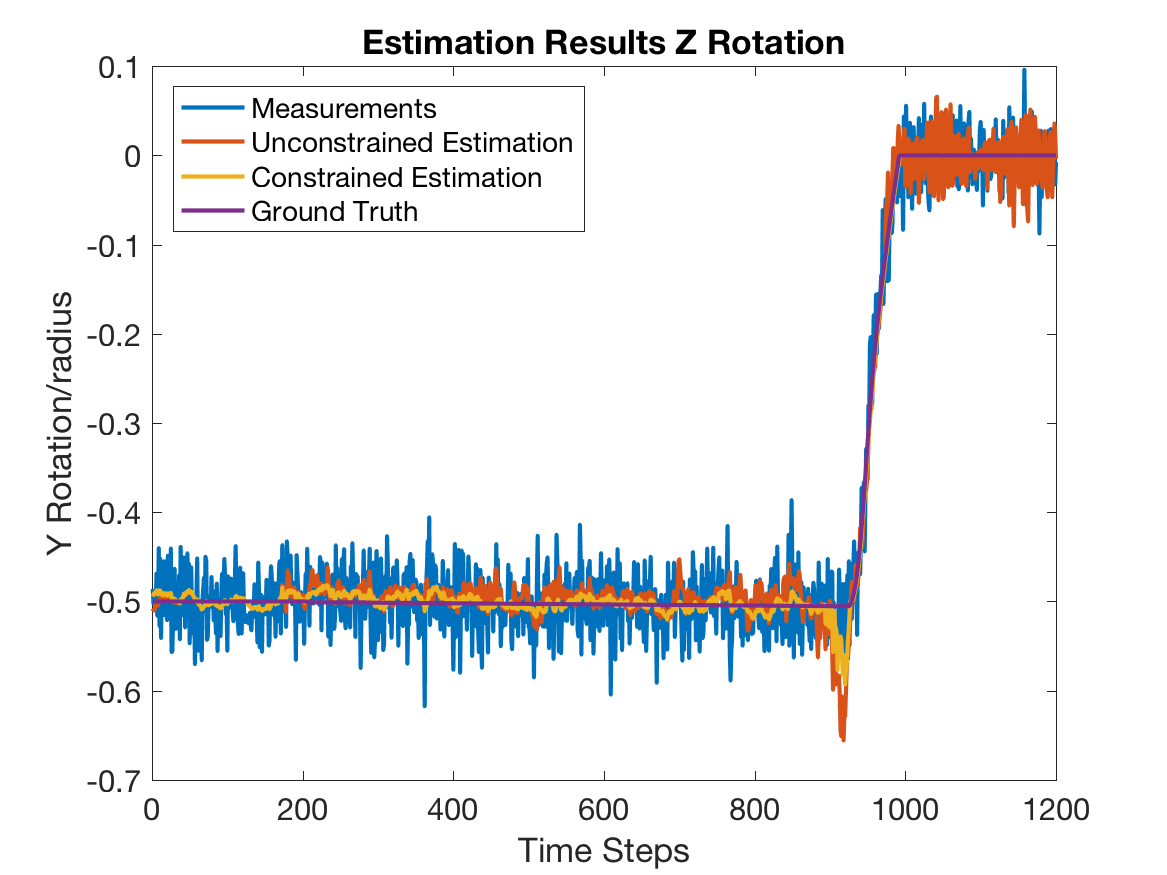}
  \caption{Estimation result for the z rotation of the object in the gripper.}
  \label{fig:estimation-results-gripper-z}
\end{subfigure}
\hfill\begin{subfigure}{.33\textwidth}
  \centering
  \includegraphics[width=.95\linewidth]{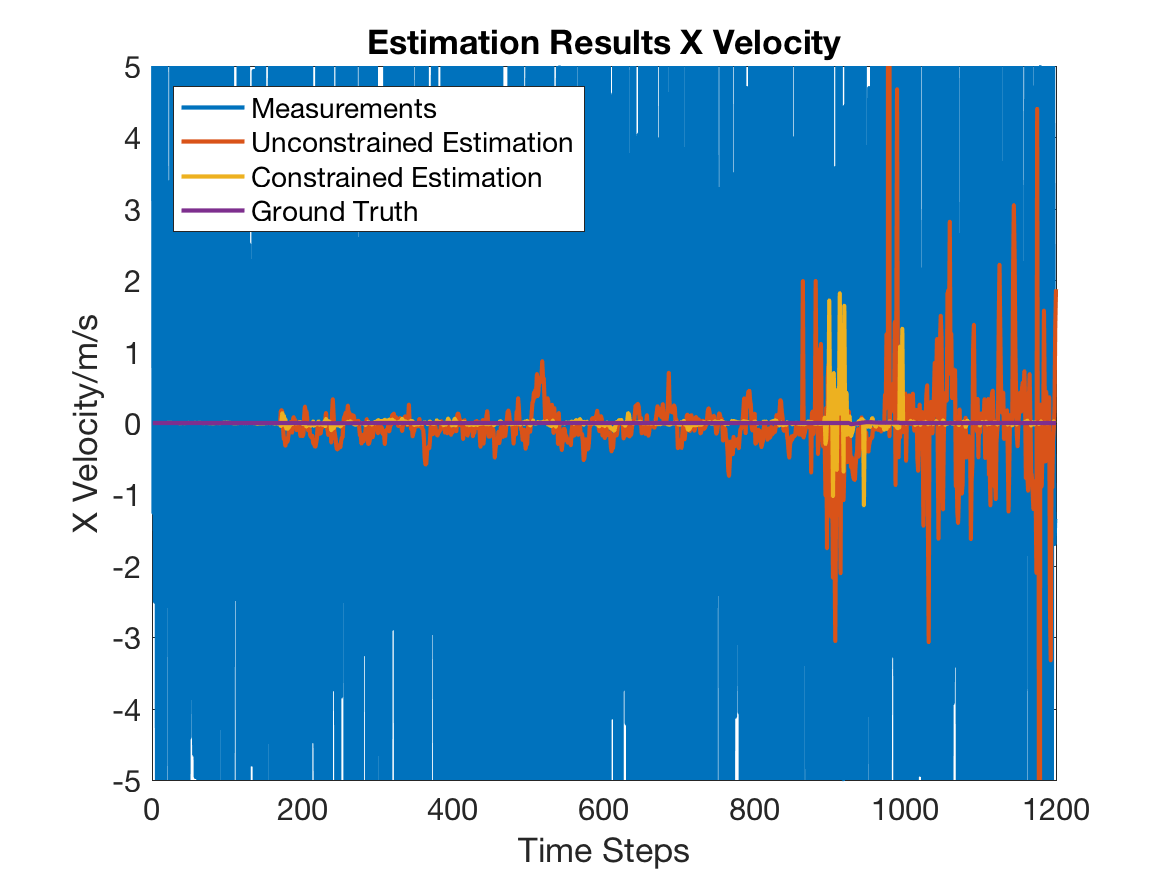}
  \caption{Estimation result for the x velocity of the object in the gripper.}
  \label{fig:estimation-results-gripper-xv}
\end{subfigure}%
\hfill\begin{subfigure}{.33\textwidth}
  \centering
  \includegraphics[width=.95\linewidth]{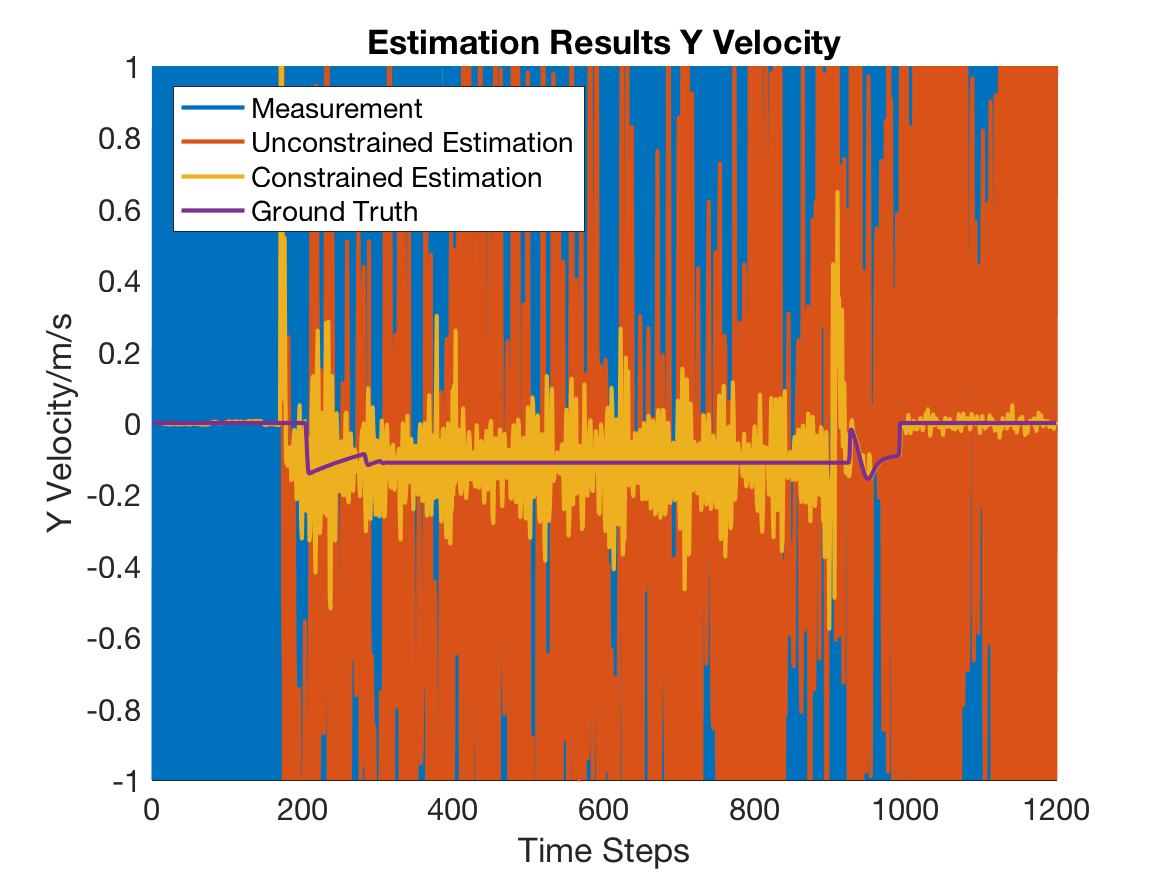}
  \caption{Estimation result for the y velocity of the object in the gripper.}
  \label{fig:estimation-results-gripper-yv}
\end{subfigure}%
\hfill\begin{subfigure}{.33\textwidth}
  \centering
  \includegraphics[width=.95\linewidth]{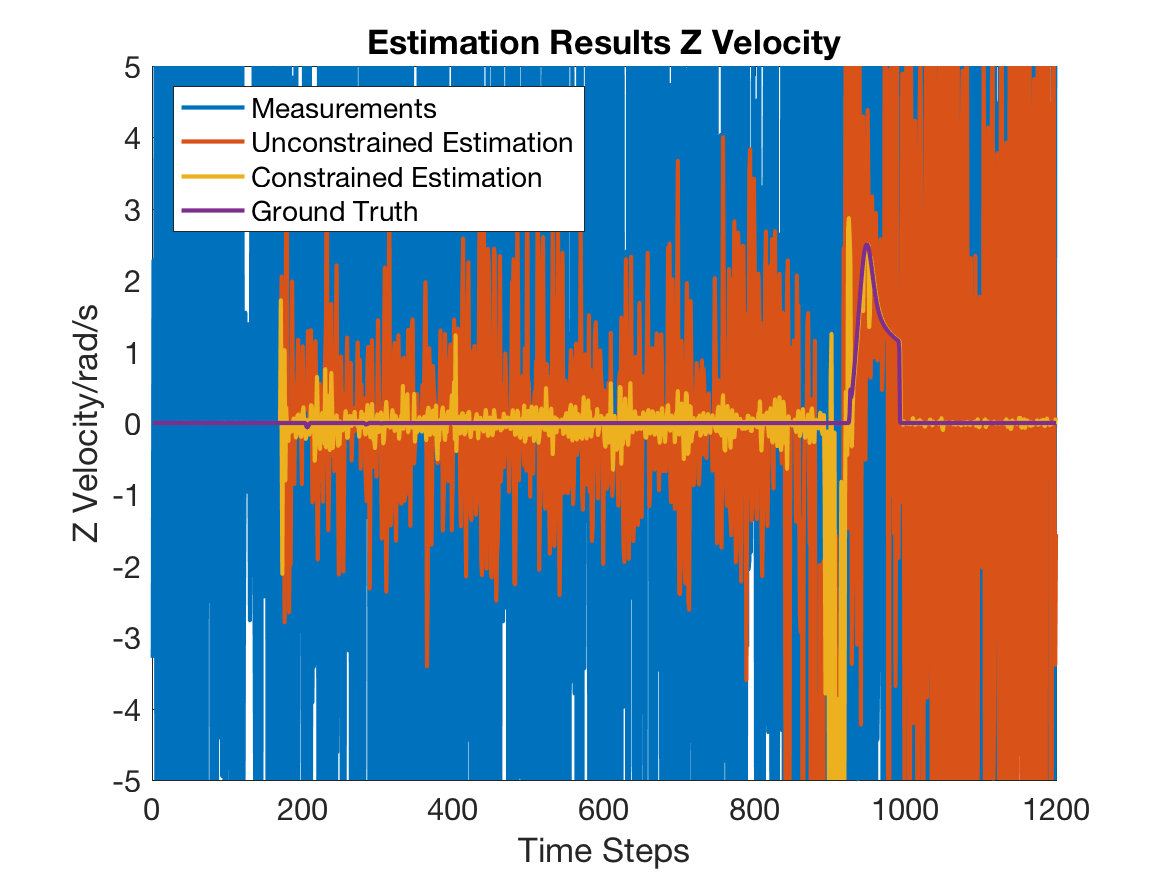}
  \caption{Estimation result for the z velocity of the object in the gripper.}
  \label{fig:estimation-results-gripper-zv}
\end{subfigure} 
\caption{Estimation results for the second set of simulation experiments.}
\label{fig:estimation-results-gripper}
\end{figure*}

Similarly, for the second set of simulation experiments, the results of both filters are shown in Figure \ref{fig:estimation-results-gripper}. The colors of the trajectories have the same meaning with those in Figure \ref{fig:estimation-results-block}. For the pose estimation results in Figure \ref{fig:estimation-results-gripper-x}, \ref{fig:estimation-results-gripper-y} and \ref{fig:estimation-results-gripper-z},  we can draw similar results with those from Figure \ref{fig:estimation-results-block-x}, \ref{fig:estimation-results-block-y} and \ref{fig:estimation-results-block-z}. Before the gripper making contact with the object at time step 200, both filters perform similarly on estimating the object's positions and orientations. The estimation accuracy of constrained contact-based RBPF is improved once the object is in contact with the upper finger at time step 200, which constrains object's translation along the y-axis. Similarly, the performance of the constrained contact-based RBPF on estimating the z rotation of the object is significantly improved once both fingers touch the object after time step 950. The covariance values of the constrained contact-based RBPF again outperforms the contact-based RBPF. Interestingly, although the object's translation along the x-axis is only constrained by friction constraints, the estimation of the constrained contact-based RBPF is improved once the object's all three vertices are in contact with the gripper after time step 1000. We also notice that the performance of the contact-based RBPF gets worse when the constraints get tighter when both fingers are in contact with the object between time steps 1000 and 1200. 

The velocity estimation results in Figure \ref{fig:estimation-results-gripper-xv}, \ref{fig:estimation-results-gripper-yv} and \ref{fig:estimation-results-gripper-zv} also emphasize our discoveries in the first set of simulation experiments, where both filters outperforms the induced velocity trajectories and the constrained contact-based RBPF outperforms the contact-based RBPF completely. 

Finally, to make a throughout comparision, we run 20 experiments of both the block experiment (Fig. \ref{fig:wall-exp-setup}) and the robotic gripper experiment (Fig. \ref{fig:gripper-exp-setup}). Additionally, we include the baseline algorithm DBC-SLAM from \cite{shuai-icra-2015}, which also incorporate a dynamic model in their filtering method. All three methods: constrained contact-based RBPF (CCRBPF) with contact-based RBPF (CRBPF) and the DBC-SLAM are compared with their averaged Root Mean Squared Error (RMSE) for estimations on both the object's poses and velocities in the table below:
\begin{table}[h!]
\centering
\begin{tabular}{|c|c|c|c|}
\hline
RMSE&CCRBPF&CRBPF&DBC-SLAM\\
\hline
Pose(Box) &$0.015\pm0.004$&$0.0314\pm0.01$ &$0.29\pm0.3$\\
\hline
Pose(Gripper) &$0.04\pm0.01$&$0.098\pm0.12$ &$0.55\pm0.01$\\
\hline
Velocity(Box )&$0.35\pm0.08$&$2.02\pm0.49$&$2.19\pm1.03$\\
\hline
Velocity(Gripper) &$0.61\pm0.36$&$45.96\pm54.82$&$0.25\pm0.22$\\
\hline
\end{tabular}
\end{table}
\vspace{-4mm}

From the table above, we confirm the conclusion from \cite{shuai-icra-2015} that on average contact-based RBPF outperforms DBC-SLAM on estimating the pose of the objects. Additionally, we found that constrained contact-based RBPF outperforms contact-based RBPF with significant improvements on estimating the velocities of the objects. DBC-SLAM turns out to  perform the best among the three methods on estimating the velocities for the robotic gripper experiments. This is a result of the noisy introduced by the Kalman filter update to the velocity estimations of CCRBPF and CRBPF during the phase when the object is fulled grasped.

Due to more computation is needed for each time step, we found the constrained filter does need more time to compute the estimation and cannot run in real time. We hope to reduce the computation time by parallelizing the particles as discussed in \cite{ZhangThesis}.
\section{Conclusion and Future Work}
In this paper, we propose the constrained contact-based RBPF based on the contact-based RBPF proposed in \cite{shuai-thesis}. In the constrained contact-based RBPF, the constraints that are used to derive the linear transition model for the contact-based RBPF are further enforced in the form of solving a quadratic programming problem. Two sets of simulation experiments were conducted. The experiment results show that constrained contact-based RBPF outperforms contact-based RBPF when there are constraints (contacts/friction forces) in the multi-body dynamic system. We further find that as the constraints get tighter, the constrained contact-based RBPF performs better while the contact-based RBPF performs worse. The results also show that the constrained contact-based RBPF is able to infer the velocities of the objects accurately while contact-based RBPF fails at estimating the velocities. In the future, we would like to verify the constrained contact-based RBPF in physical experiments. Additionally, we will also work on improving our filter's performance, such as adjusting the weights of particles based on the correction amount.

\bibliographystyle{IEEEtran}
\bibliography{mybibfile}

\end{document}